\documentclass[sn-mathphys,Numbered]{sn-jnl}%

\usepackage{graphicx}%
\usepackage{multirow}%
\usepackage{amsmath,amssymb,amsfonts}%
\usepackage{amsthm}%
\usepackage{mathrsfs}%
\usepackage[title]{appendix}%
\usepackage{xcolor}%
\usepackage{textcomp}%
\usepackage{manyfoot}%
\usepackage{booktabs}%
\usepackage{algorithm}%
\usepackage{algorithmicx}%
\usepackage{algpseudocode}%
\usepackage{listings}%
\usepackage{caption}
\usepackage{float}
\usepackage{bbm}

\theoremstyle{thmstyleone}%

\theoremstyle{thmstyletwo}%

\theoremstyle{thmstylethree}%
\usepackage{pdfpages}

\begin{document}

\title[Article Title]{Reinforcement Learning Tutor Better Supported Lower Performers in a Math Task\vspace{1em}}

\author[2]{ \sur{Sherry Ruan}}\email{sherry.s.ruan@gmail.com}
\equalcont{}

\author[1]{ \sur{Allen Nie}}\email{anie@stanford.edu}
\equalcont{These authors contributed equally to this work.}

\author[2]{\sur{William Steenbergen}}\email{w.a.t.steenbergen@gmail.com}

\author[2]{ \sur{Jiayu He}}\email{jiayuhe@alumni.stanford.edu}

\author[2]{ \sur{JQ Zhang }}\email{jiequanz@stanford.edu}
\author[1]{\sur{Meng Guo}}\email{mguo19@stanford.edu}
\author[2]{ \sur{Yao Liu}}\email{yao.liu.chn@gmail.com}
\author[1]{\sur{Kyle Dang Nguyen}}\email{kylen@stanford.edu}
\author[1]{ \sur{Catherine Y Wang}}\email{cyw339@stanford.edu}

\author[2]{ \sur{Rui Ying}}\email{ruiying@stanford.edu}
\author[1]{\sur{James A Landay}}\email{landay@stanford.edu}
\author*[1]{ \sur{Emma Brunskill}}\email{ebrun@cs.stanford.edu}

\affil*[1]{\orgdiv{Computer Science}, \orgname{Stanford University}, \orgaddress{, \city{Stanford}, \postcode{94305}, \state{CA}, \country{USA}}}

\affil[2]{\orgdiv{Work done while at Stanford University}, \orgaddress{\street{Stanford}, \city{Stanford}, \postcode{}94305, \state{CA}, \country{USA}}}

\abstract{Resource limitations make it hard to provide all students with one of the most effective educational interventions: personalized instruction.  Reinforcement learning could be a key tool to reduce the development cost and improve the effectiveness of intelligent tutoring software that aims to provide the right support, at the right time, to a student. Here we illustrate that deep reinforcement learning can be used to provide adaptive pedagogical support to students learning about the concept of volume in a narrative storyline software. Using explainable artificial intelligence tools, we extracted interpretable insights about the pedagogical policy learned and demonstrated that the resulting policy had similar performance in a different student population.  Most importantly, in both studies, the reinforcement-learning narrative system had the largest benefit for those students with the lowest initial pretest scores, suggesting the opportunity for AI to adapt and provide support for those most in need. }


\maketitle

\section{Introduction}\label{sec1}

Many children fail basic reading and math standards, and the number of such students has greatly increased during the COVID-19 pandemic.  
One-on-one human tutoring can be highly effective~\cite{nickow2020impressive}, in part because it enables students to receive personalized, differentiated instruction, but it is often prohibitively expensive. Educational software aims to provide some of this personalized instruction  at scale, but can still be costly and slow to build. 
Reinforcement learning (RL) could reduce the cost of developing effective learning technology by automating the process of specifying how best to support a student through their learning journey.  
RL algorithms learn from data to choose an intervention (such as a hint), given the current context (such as an estimate of a student's knowledge) to maximize the expected value of some desirable outcome, such as test scores.
Preliminary work on using RL for improving educational software has enabled encouraging gains on learning outcomes~\cite{mandel2014offline,chi2011empirically,park2019model,bassen2020reinforcement} or student persistence~\cite{mandel2014offline,bassen2020reinforcement}. Such systems have been limited to selecting among practice items. It is unknown if reinforcement learning could be used to automatically tune and optimize broader types of learning systems, such as the pedagogical feedback provided in a narrative environment, and do so in a way that is interpretable and robust. 

To address this, we created a narrative-based adaptive pedagogical-supported educational software to support math concept learning for students roughly ages 9-12 and used reinforcement learning to adaptively (machine) learn the responses to provide support for student learning. Recent advances in explainability methods for deep neural networks have made it possible to use advanced tools for modeling without sacrificing interpretability.  We used these methods to help understand if and how the system is learning to differentiate in order to optimize desired outcomes. An additional key consideration is whether the learned pedagogical support would generalize to a different student community, as all schools may not be able to support online adaptive RL systems. We tested if the decision policies learned in the first study could be used in a different population of students that was a  more geographically diverse population with a lower household income distribution. In both studies, students with the lowest pretest scores improved using our RL-powered narrative AI system, and more than compared to students using a baseline system. This highlights the potential for reinforcement learning  to tune educational software parameters to enhance effectiveness, in a way that is interpretable, transfers to other populations, and can help those most in need of support. 

\section{Related Work: Reinforcement Learning for Student Learning}

Reinforcement learning has seen impressive successes in areas like robotics~\cite{levine2016end} and game playing~\cite{silver2018general}. The goal of a reinforcement learning algorithm is to compute a strategy (referred to as a “policy”) that specifies the intervention (such as a pedagogical activity) to choose in a particular context (e.g., a learner’s knowledge state and frustration level), in a way that is expected to maximize desired outcomes (e.g., test scores, engagement, retention). A key challenge is that the  parameters governing the process by which contexts evolve, and outcomes occur, are unknown in advance. Instead, an algorithm must learn from experience by analyzing actual decisions made and their outcomes, a strategy with high expected outcomes.

In the context of education, there have been some promising results that reinforcement learning can improve word acquisition of preschoolers interacting with a social robot~\cite{park2019model},  the persistence of learners during a fractions game~\cite{mandel2014offline}, the performance of college students learning introductory physics~\cite{chi2011empirically}, undergraduates learning discrete  mathematics~\cite{zhou2019hierarchical}, 
and the outcomes and efficiency of working adults learning linear algebra~\cite{bassen2020reinforcement}. However, in other settings, there has been little benefit over a reasonable control condition~\cite{rowe2015improving,doroudi2019s}. 
More broadly, work on intelligent tutoring systems and computer-assisted learning suggests that personalized feedback and support in educational software can be an effective way to support student learning~\cite{corbett2001cognitive,beal2010evaluation,vanlehn2011relative}, but most prior work has focused on software designed to be used in the classroom where there are additional mechanisms to keep students' attention. %

We hypothesize that reinforcement learning may be particularly beneficial when learning is happening out of the classroom, or motivation and engagement are particularly critical, or in less traditional curricula that move towards different forms of instruction rather than lecture and practice. Learning sciences offer less guidance about how to best support students in these settings. Yet, such educational settings are likely to be increasingly important in the future, both due to immediate challenges due to the covid-19 pandemic and aftermath, as well as due to the types of skills needed for success in the 21st century. Reinforcement learning may inform  data-driven instruction for such settings, and we focus our attention on learners outside the classroom in this work. 

As another contrast between our focus and prior related work, in the context of education, it is both important and of interest to understand what the algorithm learns to do: what personalized decisions are made for different contexts and individuals, and who is most helped by the algorithm. Such issues have been historically largely unstudied in the reinforcement learning research community, with some notable exceptions (e.g.~\cite{shen2016reinforcement,zhou2022leveraging}),  but are an important part of our current work. 

\section{Interface Design}\label{sec2}

Learning science principles can often be too broad to inform the specific design decisions needed to create engaging, effective educational software. For example, a 
narrative-based, chatbot-supported\footnote{Note that our work was conducted before the launch of ChatGPT in November 2022.} educational interface can lead to significant learning and engagement gains over a no-narrative, no-chatbot variant~\cite{woz}, but doing so well is subtle. Here the effective chat-based tutoring system actually used humans to act as chatbots, in a wizard-of-oz style study. In contrast, a different  narrative-based system with standard step-by-step hints (which are common in intelligent tutoring systems) 
provided no benefit over the no-narrative, no-hint control condition~\cite{woz}. 

\begin{figure}[!th]%
    \centering
    \includegraphics[width=\linewidth]{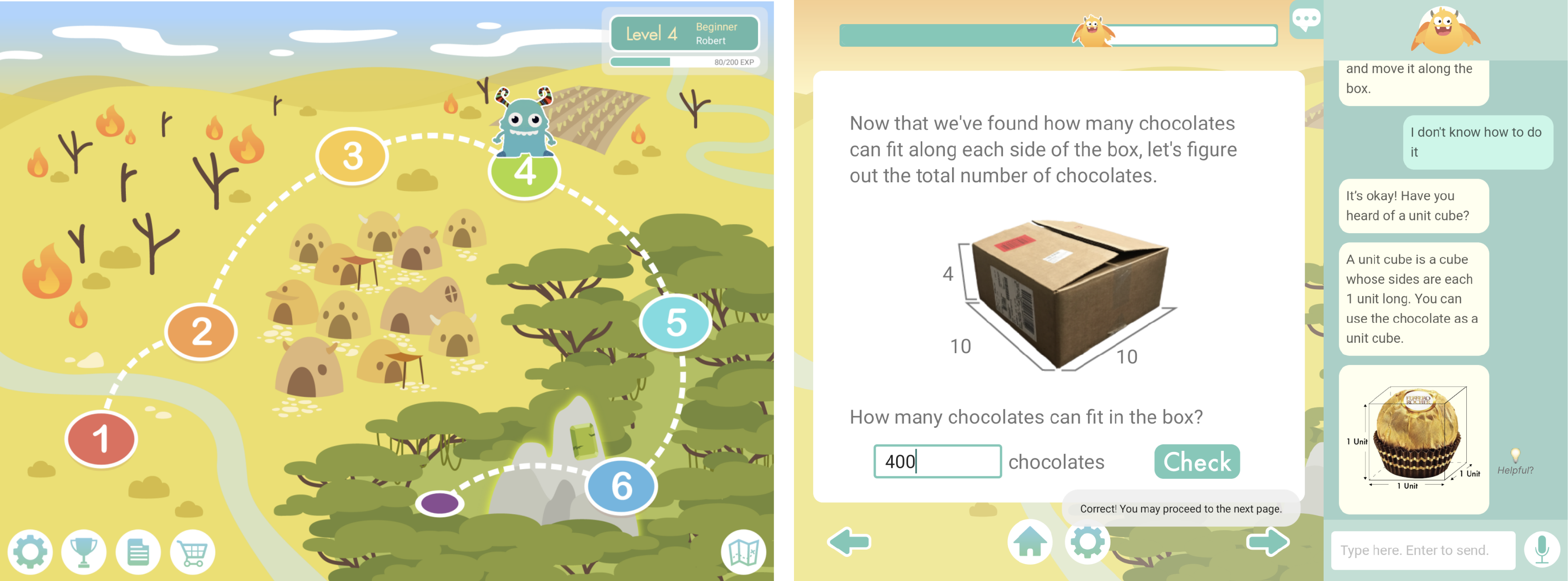} 
    \caption{\textbf{Tutoring AI Guide Interface}: A child solves a math problem while interacting with the AI-driven tutoring guide. The child can click on the “helpful?” button if they consider the AI tutor’s response to be helpful. The child can also click on “I want to stop playing” to quit the activity at any time.}
    \label{rlbot:fig:chatbot_interface}
\end{figure}

RL has the potential to be particularly helpful in such situations where personalization may be key. In this work, we used an informal online learning environment to teach students about the concept of volume. Learning tasks in this system are embedded in a narrative storyline. In response to student input, a companion AI tutor selects among four common pedagogical strategies: providing direct hints, generic encouragement, and guided prompts that scaffold the student (e.g., "Have you heard of a unit cube?"), or passive positive acknowledgment (emoticon smiley face).  Figure \ref{rlbot:fig:chatbot_interface} shows a screenshot of the software used. 

\section{Approach}

\subsection{Feature Space}

Due to past success in RL systems for adult learning~\cite{chi2011empirically,bassen2020reinforcement}, we use a small set of features, specifically 
an eight-dimensional state space, described in detail below. The observation vectors were normalized element-wise before being used for training and prediction. 
Grade, pre-score, and anxiety score are static variables. Other variables are affected by the actions the policy takes and change as the child is solving each step of the task. 

 \begin{itemize}
 \setlength{\itemindent}{.2in}
     \item Grade: the elementary school grade a child is in, ranging from 3--5. 
     \item Pre-score: the score a child receives for the pre-test, ranging from 0--8.
     \item Step: the step of the task a child is in, ranging from 1--6.
     \item Failed attempts: the number of failed attempts made by the child in the current step. It is a non-negative integer.
     \item NLP positive score: a score that reflects the positive sentiment in the last sentence mentioned by the child. It is a float ranging from 0--1. An automatic sentiment analysis tool from NLTK \cite{nltk} is used to calculate this.
     \item NLP negative score: a score that reflects the negative sentiment in the last sentence mentioned by the child. It is a float ranging from 0--1. An automatic sentiment analysis tool from NLTK \cite{nltk} is used to calculate this.
     \item NLP help score: a score that reflects the extent to which the child asks for help in the message sent. It is a float ranging from 0--1 and calculated as the semantic similarity between the child's message and ``help''.
     \item Anxiety score: the score of the math anxiety test \cite{math_anxiety} that the child takes prior to beginning the activity.
 \end{itemize}

\subsection{AI Guide RL Policy Learning}

\subsubsection*{The simulation phase}

 RL algorithms were run on a simulator before any real-world experiments were done to get an initial estimate of the performance and test the algorithm's potential. The simulator models children with various characteristics and their interactions with the math problem and agent, 
  and is built  with transition matrices of much higher dimensionality than the state space passed into the algorithms, to ensure that the simulator is challenging. 
 We select the hyperparameters of our RL policy based on this simulator.
 These early simulations informed our choice of a small function model for use in our later experiments. For example, we explored various multiple policy architectures and converged on 2 hidden layers,  since in our simulations the parameters for a small instructional model could be learned within a couple of hundred simulated students.

\subsubsection*{Online learning Phase}

\begin{figure}[!ht]
    \centering
    \includegraphics[width=0.8\textwidth]{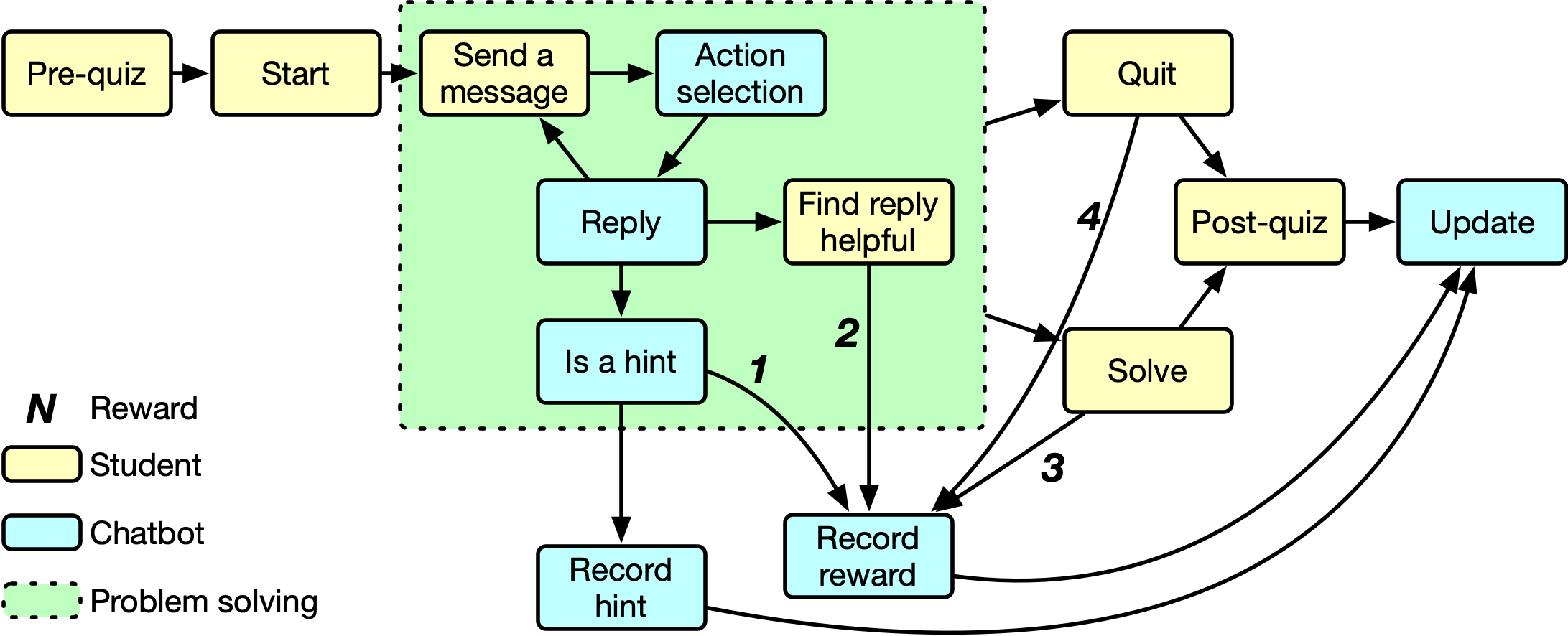}
    \caption{The interaction between user and RL AI guide. The RL AI guide selects one of four actions and replies to the user once a message is received. The reward function is updated both during the interaction and after the child completes the post-quiz. Rewards 1--4 correspond to the reward functions described in Section 4.2 (Reward Function). The RL AI guide performs an update after every five children.}
    \label{rlbot:fig:rl_diagram}
\end{figure}

Throughout the math-learning activity, children have access to an AI guide on a side panel that provides encouragement, hints, and companionship. The goal is for the AI guide to provide additional engagement with the math activity and provide adaptive support that facilitates learning gains. The AI guide takes on the persona of the monster that children select in the fantasy-based narrative. Before entering the math learning activity, children are brought through a short tutorial in which they communicate with the AI guide, which introduces itself and asks about the children. This tutorial serves to familiarize the children with the AI guide interface and build social rapport between the AI guide and the children. We provide a workflow in Figure~\ref{rlbot:fig:rl_diagram}.

The RL decision policy takes in a vector describing features of the learner state and outputs a particular support type (of the 4 options) to provide. 
The RL algorithm aims to learn an automated decision policy to maximize the expected reward function, which should capture the key desired outcomes. We specify the reward when a student $j$ finishes as:
\[
R_j = \sum_{i=1}^8 [\max (0, post_{ij} - pre_{ij})] - \lambda * n_{hj}  + \beta n_{uj} + \mathbbm{1}(quit_j),
\]
$(\lambda=0.013,\beta=0.1)$, where the first term is the sum over items of the $j$-th student's clipped learning gain from pretest to post-test %
on item $i$ of the assessment, the second term is a tiny penalty on the number of hints $n_{hj}$ given by the system to the student (since too many hints may reduce learning), the third term provides a small bonus for the number of times $n_{uj}$ child $j$ marked an AI guide reply as helpful, and the last term  $\mathbbm{1}(quit_j)=-8$ is a penalty if the learner quits  before completing the task. The proximal policy optimization (PPO) algorithm~\citep{schulman2017proximal} was used to learn the decision policy to optimize the expected reward. 

The policy architecture is stochastic. The hyper-parameter used in the online study was $\epsilon=0.2$. Both the policy neural network and value function neural network had two hidden layers with 16 nodes and a tanh activation function. We used an Adam optimizer with a learning rate of 0.0025 for both.
The RL policy is implemented with the RLGraph package for this \cite{Schaarschmidt2019}. This optimization method was chosen as it has shown potential in similar situations, for example, in \cite{bassen2020reinforcement}. 

\subsubsection*{Offline Reinforcement Learning}
\label{sec:orl}
We also performed offline reinforcement learning to extract another policy for use in a subsequent experiment. We did this for multiple reasons. First, as described later, during online reinforcement learning, the policy had not yet converged by the end of study 1, and we wanted to compare a static learned policy to a control, where the differences might be clearer. Second, we were curious whether we might extract a higher-performing decision policy using offline learning. Third, in most experimental sciences, research is hoped to provide findings that generalize beyond the specific research setting. Such generalizability is also of key interest in machine learning. Therefore an important open issue is whether  automated pedagogical strategies obtained using reinforcement learning in one setting will transfer to similar settings.  

We used offline reinforcement learning policy evaluation to select among potential new automated instructional policies using the data gathered from online reinforcement learning (in our study 1, as we will shortly describe). We considered two sets of algorithms for training potential instructional policies. The first is behavior cloning~\cite{pomerleau1990rapidly,sammut1992learning}, a popular method for leveraging offline data to train an automated policy. Behavior cloning trains the model to imitate the probability distribution of actions that are outputted by the online policy. Note that this procedure does not return a policy that is exactly like our online policy because our online policy updates itself -- therefore, this objective trains a new policy that tries to output the probabilities outputted by an ensemble of online models at different checkpoints. Behavior cloning minimizes the following loss:
\[
\mathcal{L}_{\text{BC}}(\theta, \mathcal{D}) = \mathbb{E}_{(s, a, s') \sim \mathcal{D}}[D_{\text{KL}}(\pi_\theta(s) || p(a|s))]
\]
In our setting, this can be viewed as distilling the average policy over online reinforcement learning.

The second style of algorithms we explored was offline policy gradient on the estimated performance of the trained instructional policy. This method has been used in several other offline RL optimization papers (see e.g.~\cite{metelli2018policy,liu2020off}). Here we used a weighted importance sampling (WIS) estimator to estimate the value of the policy, 
\begin{align}
   \mathcal{L}_{\text{WIS}}(\theta, \mathcal{D}) &= \frac{1}{\sum_{i=1}^{|\mathcal{D}|} \big( \prod_{t=1}^L \frac{\pi_\theta(a_t|s_t)}{p(a_t | s_t)} \big)} \sum_{i=1}^{|\mathcal{D}|} \Big( \prod_{t=1}^L \frac{\pi_\theta(a_t|s_t)}{p(a_t | s_t)} \Big) R_i \\
    & + \eta \cdot \frac{1}{\sum_{i=1}^{|\mathcal{D}|} \big( \prod_{t=1}^L \frac{\pi_\theta(a_t|s_t)}{p(a_t | s_t)} \big)^2}
\end{align}
where $R_i$ is the total reward for student $i$. This is called policy gradient via importance sampling (POIS). We also explored whether adding an effective sample size (ESS) penalty with hyperparameter $\eta$ would help -- ESS regularizes the difference between the learned policy $\pi_\theta$ and the behavior policy $p$.

We considered multiple hyperparameters for each of the two algorithm procedures (see Table~\ref{tab:hyperparam}). There are 108 hyperparameter combinations to learn our policy. We use an algorithm evaluation procedure where we partition the collected dataset into a train and validation set by randomly allocating 50\% of students into one group and the rest into another. We repeat this strategy 10 times. We use this splitted dataset to choose the best model architecture, hyperparameters, and learning objectives, similar to what has been proposed in \cite{nie2022data}. We trained our model on the training split and use weighted importance sampling (WIS) to evaluate the performance of this policy on the validation set. We apply the same learning procedure across all 10 splits and compute the average of the performances. We choose the best algorithm from the highest average performance on the validation set. We then apply this algorithm to train a policy that learns from the entire dataset.

\begin{table*}[tb!]
\centering
\begin{tabular}{@{}cc@{}}
\toprule
     & \begin{tabular}[c]{@{}c@{}}Hyperparameter Range\end{tabular} \\ \midrule
\begin{tabular}[c]{@{}c@{}}Training algorithm\end{tabular} & BC, POIS \\ \midrule
\begin{tabular}[c]{@{}c@{}}Policy network  dimension\end{tabular} & \begin{tabular}[c]{@{}c@{}} [4], [8], [16], [4,4],  \\ \text{[8,8]}, [16,16] \end{tabular}                                             \\ \midrule
\begin{tabular}[c]{@{}c@{}}Training  Epochs\end{tabular}                   & [1, 5, 10]                                                   \\ \midrule
\begin{tabular}[c]{@{}c@{}}ESS Penalty $\eta$ \end{tabular}                      & [0, 0.01, 0.05]                                                \\ 
\bottomrule
\end{tabular}
\caption{Hyperparameters considered during offline batch reinforcement learning.}
\label{tab:hyperparam}
\end{table*}

In our evaluation, the behavior cloned policy was estimated to outperform the online policy in the majority of splits. Also, a small 1-layer fully connected neural network with 4-dimensional hidden state and Gaussian error linear unit~\cite{hendrycks2016gaussian} activation function outperformed other model architectures. 

Therefore we used the distilled, behavior cloned policy in our second experiment

\section{Experimental Setups}

As a control condition, the interface included the mathematics task but had no narration and no adaptive support; similar to a mastery-style approach, students had to successfully complete one subpart before advancing. While this may seem like a weak control, a past study~\cite{woz} on teaching an elementary school mathematics task had found that a similar control condition had performed similarly to a control condition with a narrative storyline, and slightly better than a control condition with a narrative storyline and step-wise hints (which are common in tutoring software).

In study 1, we examined the speed and effectiveness of using reinforcement learning to adapt the type of AI guide feedback given to learners. Due to COVID-19 pandemic restrictions, all experiments were completed online. Subjects were randomly assigned to each condition, but with an unequal allocation-- more students were assigned to the RL condition than the control condition. In total 269 elementary school students used the reinforcement learning-narrative educational software (RL). 70 students were in the control condition.

Subjects completed an 8 item assessment and a math anxiety survey~\cite{carey2017modified}, then used the volume education software, and then completed another assessment (identical up to numerical values, and cross-randomized across students), and an engagement measure designed for studies with children~\cite{giggle_gauge}.

\begin{figure}[bt!]
  \centering
  \includegraphics[width=.45\linewidth]{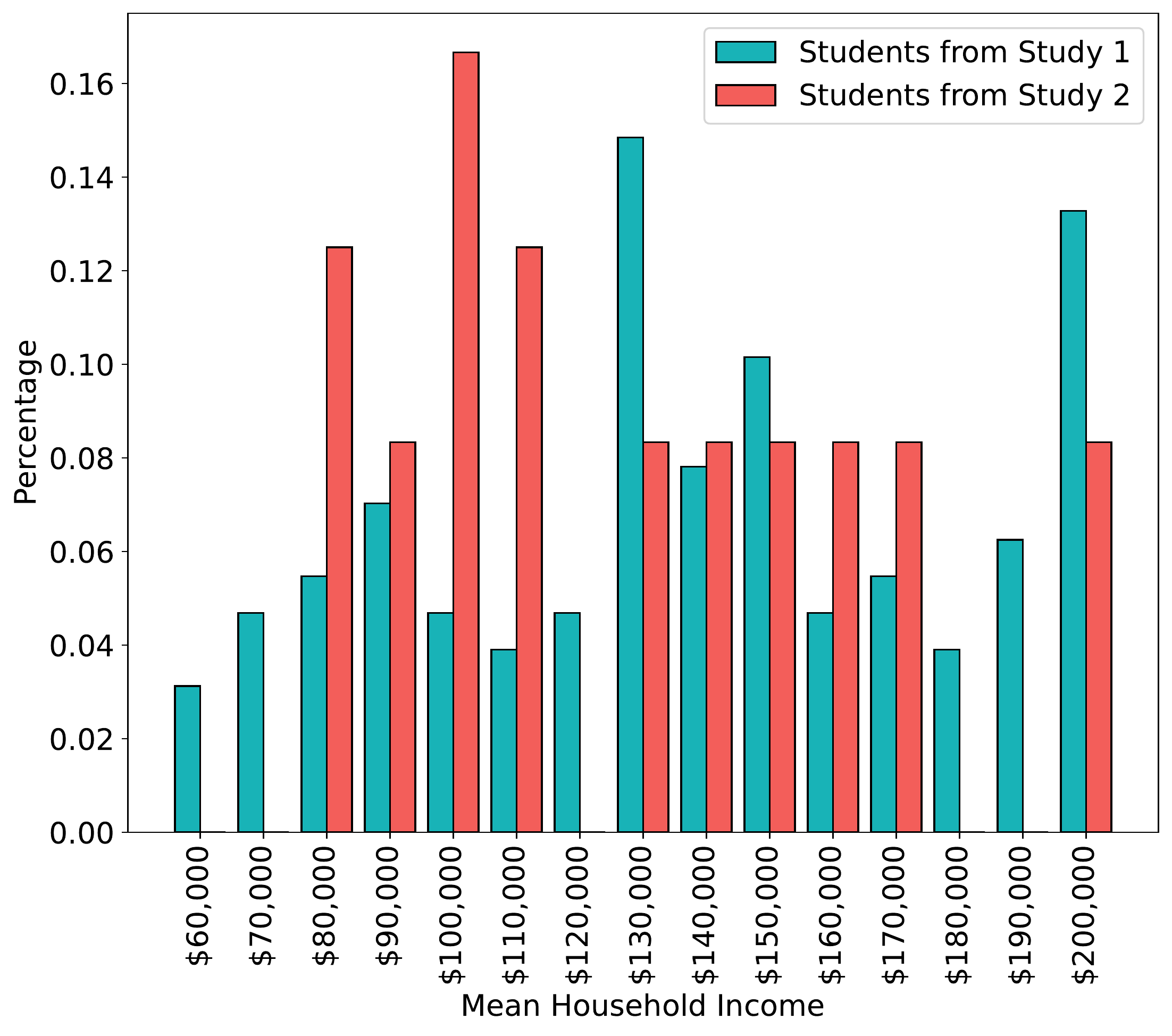}\quad
  \includegraphics[width=.45\linewidth]{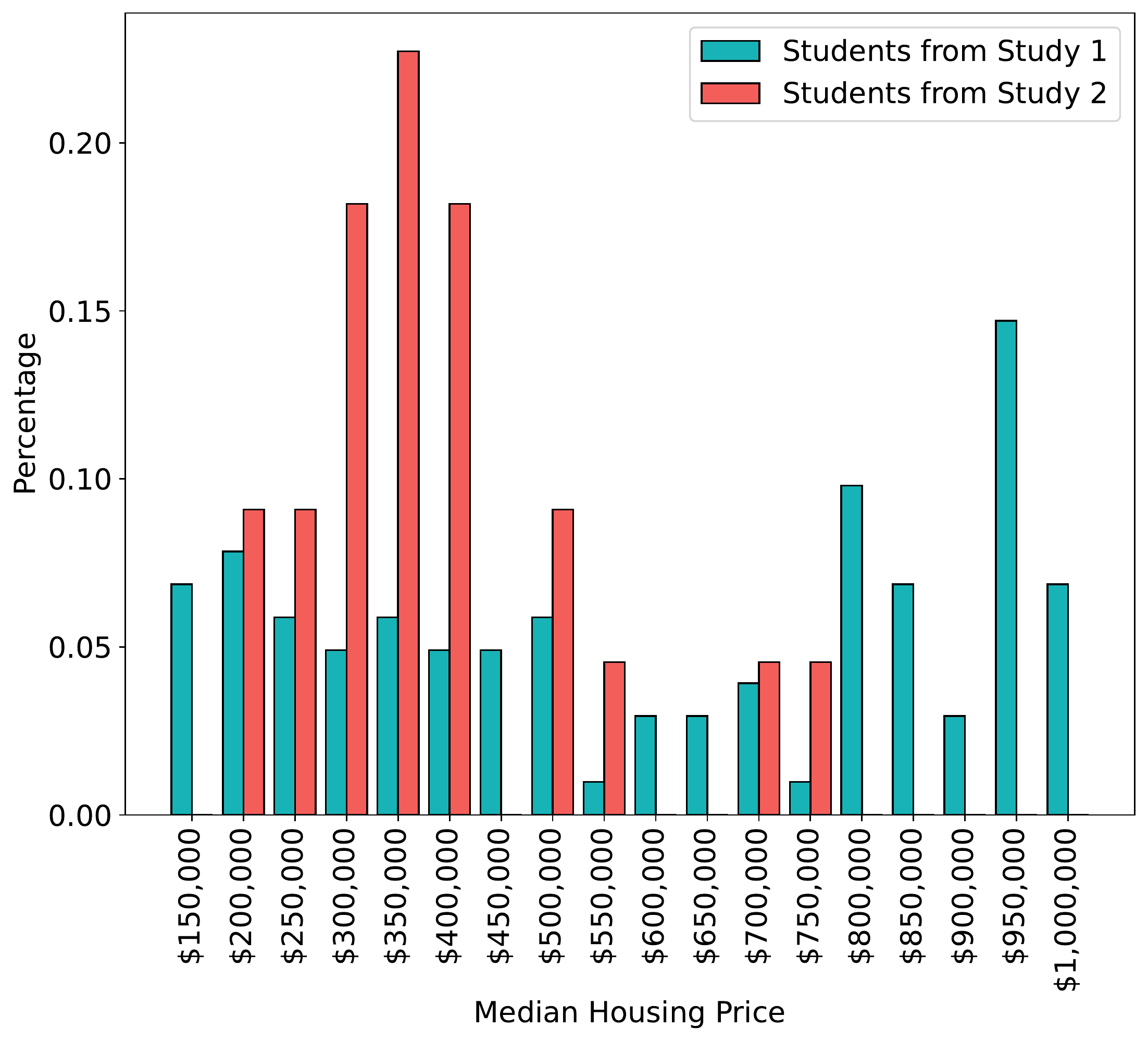}
  \caption{Distribution of household income (left) or median housing price (right) in the zip codes provided by subjects in study 1 and study 2. There significant difference in the subject pools between the two studies.}
      \label{rlbot:fig:dist_shift}
\end{figure}
In study 2 we were interested to see if the distilled behavior cloned policy learned from the online RL process (Section~\ref{sec:orl}, would transfer to a new population of subjects. 
We then conducted study 2 with a new set of subjects (37 participants used for analysis): subjects were randomized into the same control condition as study 1, or using the single distilled  RL policy. 

In study 2, we recruited a broader population more similar to that of the U.S.A.  For the original study, 113 participants out of 203 provided home zip codes. For the follow-up study, 16 participants out of 30 provided home zip codes.  For those that did not provide their home zip code, we use their school zip code. Using these zip codes, we  obtained the median housing price and mean annual household income from the fifth American Community Survey (in 2020), accessible through an API provided by the United States Census Bureau.  Figure~\ref{rlbot:fig:dist_shift} shows the difference between the student groups in study 1 and study 2. 
We conduct the Kolmogorov-Smirnov 2-sided test between student populations of two studies on these variables. For both mean annual household income ($Pr(F(x)=G(x)) = 0.02 < 0.05$) and median housing price ($Pr(F(x)=G(x)) = 0.0005 < 0.01$), we found a significant difference between two populations.
In addition, subjects were more geographically and racially diverse (see Appendix). %
In addition, study 1 was done when many more U.S.A. children attended school remotely. Thus, study 2 offers a chance to examine the generalizability of learned RL policies.

\begin{table*}[b]
\centering %
\caption{Mean (std. dev) results of children in both studies.} 
\begin{tabular}{@{}lll|ll@{}}
\toprule
& \multicolumn{2}{c}{Online RL Study 1} & \multicolumn{2}{|c}{Distilled Policy  Study 2} \\
& \textit{Control} & \textit{Narrative AI} & \textit{Control} & \textit{Narrative AI}\\
\midrule
\textbf{Number} &68  & 258  & 18 & 17 \\
\textbf{Pretest} & 5.46 (2.73) & 4.87 (2.38) & 4.06 (2.58) & 3.41 (2.31) \\
\textbf{Posttest} & 5.89 (2.57) & 5.72 (2.31) & 3.83 (2.53) & 4.12 (1.87) \\
\textbf{Improvement} & 0.44 (1.26) & 0.84 (1.89) & -0.22 (1.7) & 0.7 (2.33)\\
\midrule
\textbf{Engagement} & 3.16 (0.62) & 3.4 (0.52) & 3.17 (0.52) & 3.28 (0.57) \\
\textbf{Completion} & 92.6\% & 94.5\% & 100\% & 100\% \\
\bottomrule
\end{tabular}
\label{rlbot:tab:offline_results}
\end{table*}

\begin{figure}[tb]
\centering
 \includegraphics[height=4.0in]{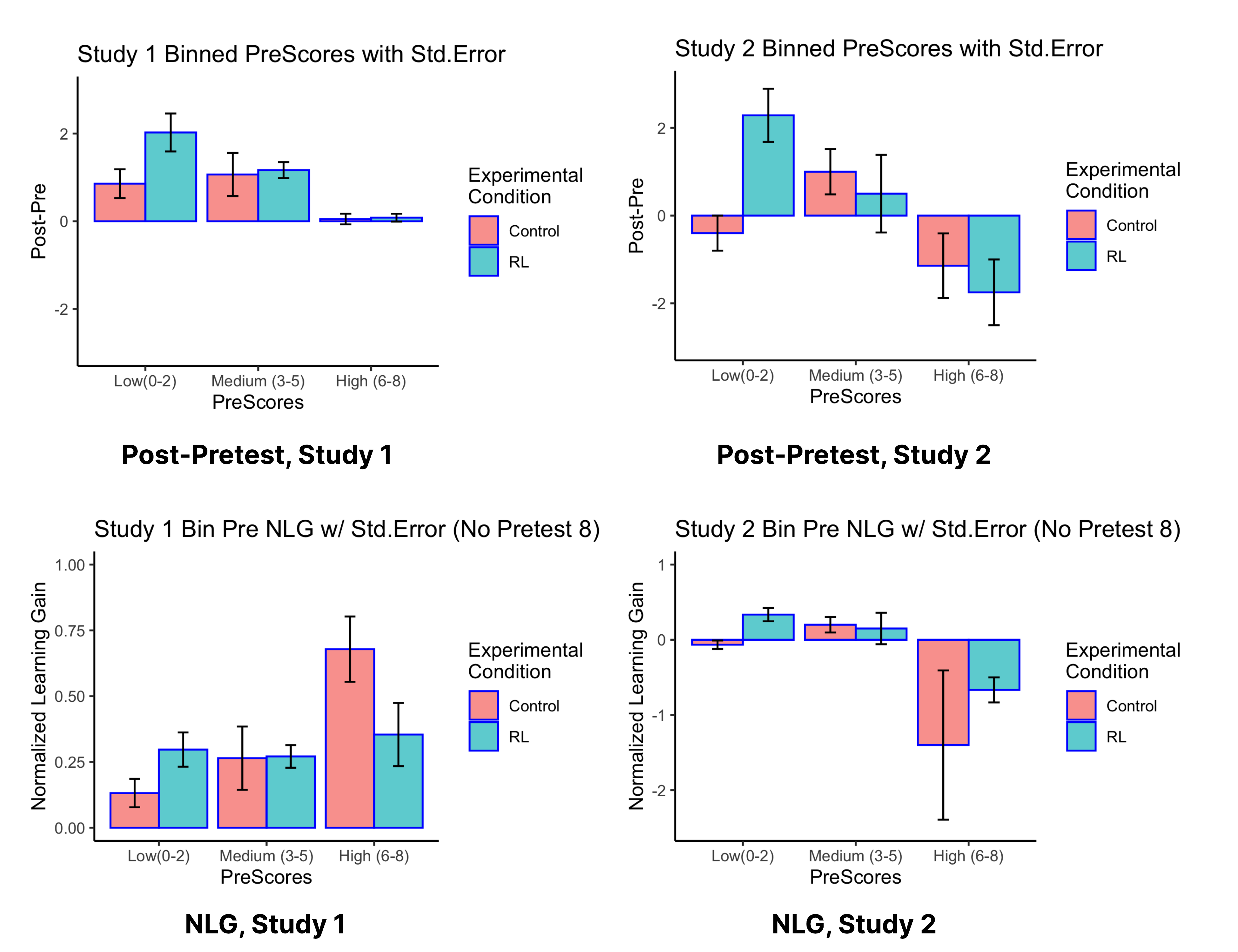}
 \caption{Top row,  Post-test - Pretest (y-axis), Bottom row, Normalized learning gain (NLG) $\frac{Post test - Pretest}{MaxScore - Pretest}$ (y-axis). Scores are clustered by subjects with low (0-2), medium (3-5), and high (6-8) initial pretest scores. Error bars show standard errors. Note the NLG (bottom row) calculations exclude students who scored 100\% on the Pretest since the NLG is not well defined.}
      \label{rlbot:fig:improve_by_pre_study}
 \end{figure}

\section{Results}\label{sec3}

Aggregate summaries are shown in Table~\ref{rlbot:tab:offline_results}. Some subjects completed the pretest or posttest twice due to a limitation in the system. We excluded these subjects from the results presented. There was no significant difference in the amount of improvement (post-test - pretest score) between the RL narrative condition and control condition (study 1: Wilcoxon rank test $W = 9632.5$, $p = 0.2$, study 2: Wilcoxon rank test $W = 185.5$, $p = 0.281$).   
 
 However, encouragingly, in both studies, there was a trend for subjects with a low initial pretest score (0-2) to have a much larger improvement between the pretest and post-test in the RL narrative condition (Figure~\ref{rlbot:fig:improve_by_pre_study}, top row). The average improvement for these students was 2.02 in study 1 (N=41), and 2.29 in study 2 (N=7), out of a total score range was (0-8). There was a significant difference in the change in scores between the RL condition and control condition in study 2 for those with low pretest scores (0-2) (Wilcoxon rank test $W=2, p=0.013$), though this difference does not persist after correcting for multiple-hypothesis testing, and all other differences for studies and pretest groups were not statistically significant under the same test.  %
 
 Engagement scores range from 1 to 4 and subjects with low initial pretest scores (0-2) also trended to having much higher engagement in the RL AI guide condition (study 1 mean engagement score 3.29 (N=40), study 2, mean engagement score 3.28 (N=7)) than in the control condition (study 1 mean engagement score 2.7 (N=14), study 2, mean engagement score 2.7 (N=5)).    
 Prior work suggests interpreting scores below 3.0 as low engagement and 3.0-3.6 as moderate engagement~\cite{giggle_gauge}.

The assessment used may be subject to ceiling effects, as a number of students did receive the maximum score (8) on either the pretest or the post-test. Though the pretest scores did not significantly differ between the two conditions, in either study, since the control pretest scores were slightly higher, ceiling effects may have impacted the control condition more.

To address this, we also repeated our analysis using normalized learning gains (NLG), $\frac{Post test - Pretest}{Maximum score - Pretest}$, which represent the fraction of improvement made by subjects, relative to the possible improvement. Note this excludes any subjects who scored the maximum score on the pretest since the NLG is not well-defined for such students. There was no significant difference between the RL narrative condition and control condition for NLG in either case (study 1, W = 4394.5, p-value = 0.6978; study 2, W = 104.5, p-value = 0.3819). 

Like for posttest - pretest, we observe larger normalized learning gains for the RL narrative condition than the control condition for initially lower performing students, in both studies (Figure~\ref{rlbot:fig:improve_by_pre_study}, bottom row). The NLG performance for students with medium pretest scores is similar in both conditions, as was also seen for such subjects' posttest minus pretest scores. The pattern for the highest performing students is slightly different than for the post-test - pretest scores but should be taken lightly: as stated, the NLG analysis ignores all students with maximum pretest scores. Note that an NLG of 75\% for the initially high-performing student group would be at most a $2*0.75=1.5$ post-test - pretest improvement (since 2 is the largest possible gain, if the student scored 6 on the pretest, and it is lower if the student scored 7), whereas a 30\% improvement for the initially low performing student group is at least a gain of $6*0.3=1.8$ on the post-test - pretest (since $MaxScore - Pretest \geq 6$ for such subjects). 

Together these analyses encouragingly suggest that the RL narrative condition trends to provide a bigger benefit to initially lower-performing students than the control condition. 
We now provide some additional analyses into the RL process and the potential mechanisms underlying this difference. 

\subsection{RL Online Learning}

In study 1, the RL agent updated the AI guide pedagogical policy over subjects, but during the 28 policy updates (after 10 subjects each), we observed significant variability, and the performance had not converged. 

We hypothesize this may be due to several factors. Likely most importantly, we saw a significant variation in the pretest scores of subjects over time. This may be in part because we performed rolling recruitment, adding additional recruitment sources during the study, which likely caused some shift in the distribution of the underlying students. In addition, the natural variation across third to fifth-graders and student background skills means that across small sets (such as the 10 trajectories used each round for PPO), it is quite possible to have a substantial difference in the pretest scores of those subjects. If any of the students are already at or near the ceiling of the pretest scores, there will be almost no potential room for improvement for the RL policy. Indeed there may be some natural regression to the mean, which means that an RL policy that looked promising in prior rounds for related states, may now look worse (depending on the particular generalization). Even without this potentially shifting population, ten trajectories (subjects) is a small size to average over when performing policy updates, so the gradient may be quite noisy. This suggests that performing stratification and trying to ensure a stable distribution of initial start states over participants might lead to faster convergence and better results. 

However, despite this, through training, subjects in the AI guide condition consistently match or exceed the average performance of those in the control condition.

\begin{figure}[tb]
\centering
 \begin{minipage}{0.45\columnwidth}
  \centering
    \includegraphics[height=1.6in]{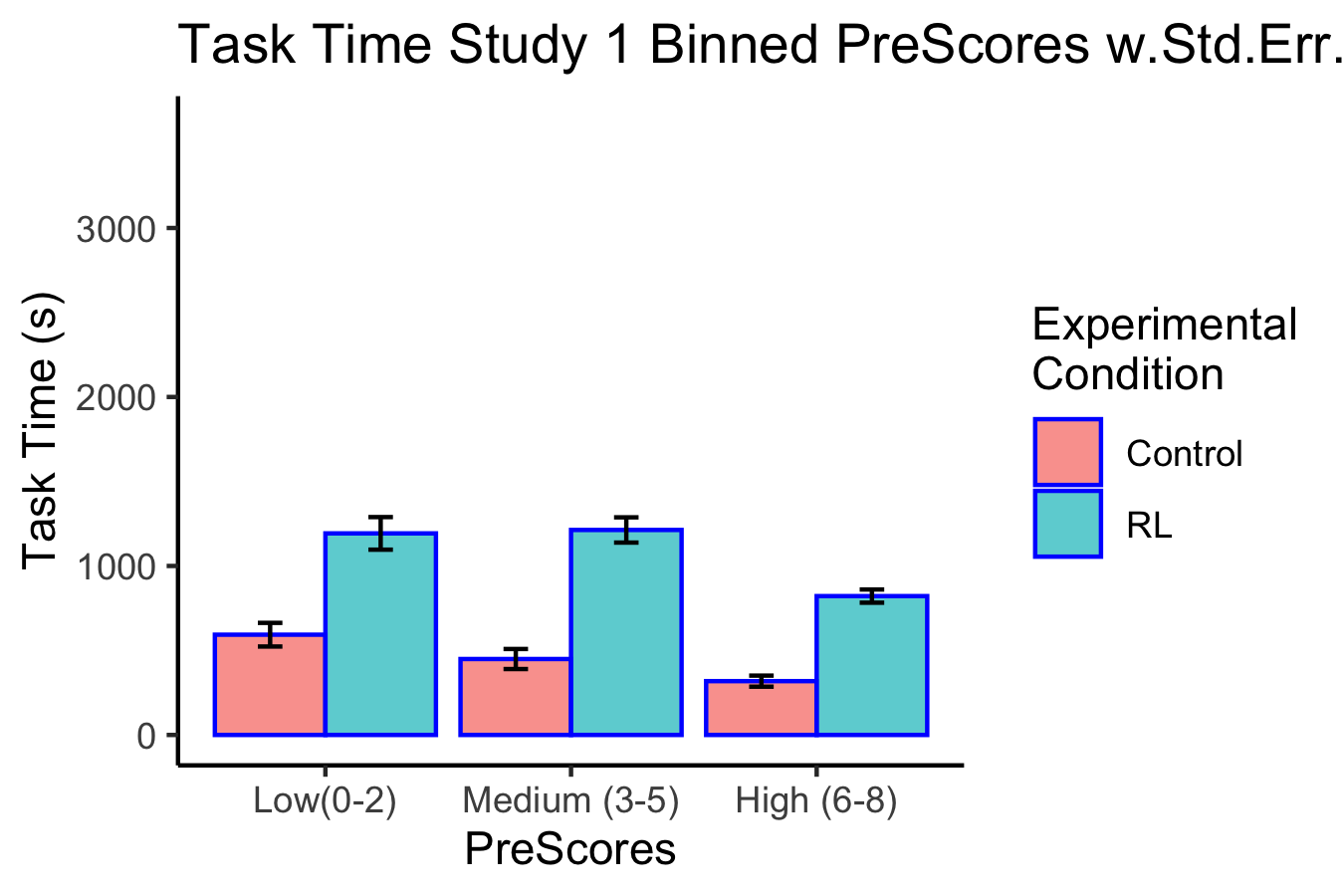}
 \end{minipage} \hfill
  \begin{minipage}{0.45\columnwidth}
\centering
    \includegraphics[height=1.6in]{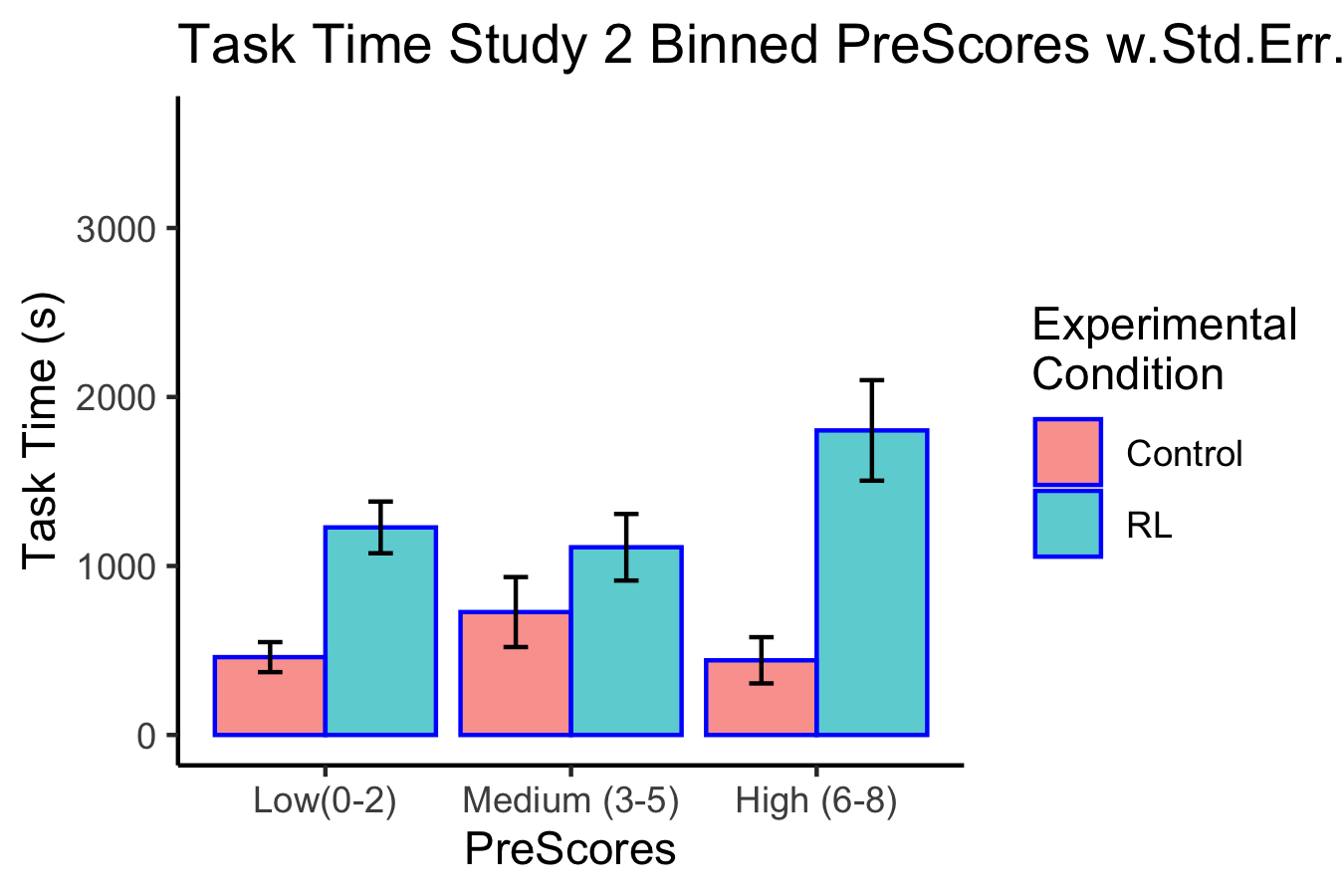}
 \end{minipage}
 \caption{Time on task (sec) (y-axis) by low (0-2), medium (3-5), and high (6-8) initial pretest scores. Error bars show standard errors. Students whose time on task exceeded 90 minutes (8 students) were excluded from the analysis since it was likely such students might have taken significant breaks.}
      \label{rlbot:fig:time_on_task}
 \end{figure}

\subsection{Investigating Other Explanations for the Benefit to Low Pretest Subjects}
A natural question is what is the mechanism behind the improved performance of subjects in the 
RL narrative condition over those in the control condition, for subjects with initially low pretest scores, and whether this could be due to factors beyond the RL-narration itself.

One potential hypothesis is that there were additional differences between the two conditions. Indeed, on average, subjects spend longer on the RL narrative condition task than in the control condition. As Figure~\ref{rlbot:fig:time_on_task} shows\footnote{We excluded individuals who took longer than 90 minutes on the task in this figure, since such subjects are likely to have taken breaks. All individuals who took at least 90 minutes took over 2 hours, and there were 8 such individuals excluded using this restriction.}, this was consistent for students across all three groups of pretest performance, and the difference in time spent between the two conditions was largely similar for all three groups. However, only the students in the low pretest group seemed to have a significant benefit from the RL condition. It seems unlikely that time on task is the primary reason for improved performance in the RL narrative condition. 

The study was conducted remotely, and a prescreening call was done with a guardian of each child participating to discuss the study, emphasize the child should do the task without assistance, and verify the child would be participating. However, it is still possible that guardians helped the children in some cases. It seems unlikely that for children with low pretest scores, guardians helped them more if the child was in the RL condition than if they were in the control condition. Indeed the control condition offered less support and hints than the RL narrative condition, so the opposite seems more likely to be true. One potential exception is that the RL narrative condition involved a storyline, and while unlikely, depending on the subject's reading skills, it is possible that the guardian would have helped the subject to understand the text.

\begin{figure}[tb!]
\centering
    \includegraphics[width=0.9\textwidth]{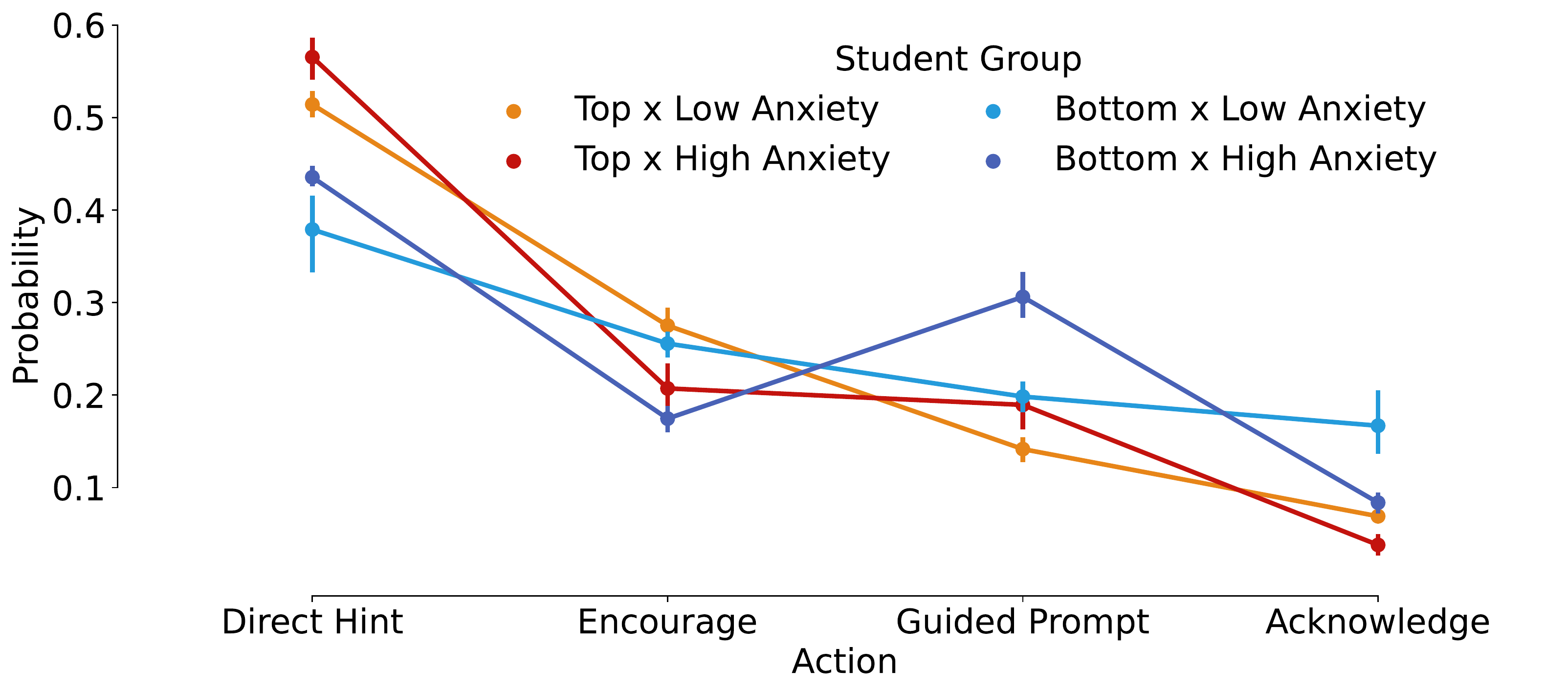}
  \caption{The y-axis shows the probability of choosing the first action  for each group of subjects, based on their pretest scores (Bottom (0-2), Top (6-8)), and math anxiety level (Low (9-13, corresponding to the bottom 25\% percentile), and High (22-45, corresponding to the top 25\% percentile)).  Error bar shows 95\% CI.}
      \label{rlbot:fig:first_step}
 \end{figure}

\subsection{Integrated Gradient Analysis of Policy on Feature Space}
A natural question is whether benefits to subjects with low pretest scores may derive from the personalization capacity of the RL instructional policy.  Indeed a key benefit of using RL to select activities is its potential to differentiate instruction if doing so is estimated to improve outcomes. Therefore it is of interest to evaluate what differentiation, if any, is done by the RL AI guide policy. However, 
most popular RL algorithms, including PPO, which we use here, use complex function approximators that are hard to interpret. Therefore we  use a method in explainable machine learning, integrated gradient~\cite{sundararajan2017axiomatic}, to decompose the multi-decision output of the RL policy used in study 2 into a linear additive sum of attribution for each input context feature.

 Table~\ref{rlbot:tab:feature_interp} shows that the feature importances computed for the policy selected from offline RL and deployed in the RL condition. Recall there are three primary categories of features used to select pedagogical strategies: static features of the learner, features about the stage of the learning activity, and features about the learner’s interaction and performance during learning. 
 
\begin{table*}[ht]
\centering
\caption{Feature importance calculated by the integrated gradient method. Numbers represent how on average, the feature (with its original value) will positively or negatively contribute to how our RL policy decides to increase or decrease the probability of choosing an action for the current student.}
\vspace{1mm}
\begin{tabular}{lccc}
\toprule
\textbf{RL Policy Action}  & \begin{tabular}[c]{@{}c@{}}\textit{Pre-test}\\ \textit{Score}\end{tabular}  & \begin{tabular}[c]{@{}c@{}}\textit{Math} \\ \textit{Anxiety}\end{tabular} & \begin{tabular}[c]{@{}c@{}}\textit{Other} \\ \textit{Features}\end{tabular} \\
\midrule
\textbf{Pr(Direct Hint)} & +7.5\% & +8.8\% & -0.4\% \\
\textbf{Pr(Acknowledgment)} & -7.5\% & -7.4\% & +0.6\% \\
\textbf{Pr(Encouragement)} & +4.1\% & -8.2\% & +0.4\% \\
\textbf{Pr(Guided Prompt)} &  -4.2\% & +6.8\% & -0.7\% \\
\bottomrule
\end{tabular}
\label{rlbot:tab:feature_interp}
\end{table*}

This analysis selected student’s pretest score and their math anxiety score as the most influential contextual features on the AI guide's chosen response.  Other student features had little to no effect.  
Figure~\ref{rlbot:fig:first_step} shows the probability of assigning actions for students from our distilled policy. 

Students with higher pretest scores were more likely to receive  direct hints: such students may %
require less of the productive struggle needed to learn new mathematics. %
Students with lower pretest scores may need more engaged practice, but those with high math anxiety may also perceive math as more effortful~\cite{choe2019calculated}.  Increasing the use of guided prompts may help support such students, as we observe in the policy instructional selections for low-performing, higher math anxiety students.  These observed interactions between the multiple features describing student and context, and pedagogy choices, could inform expert analysis and support future hypothesis generation for learning sciences.

\section{Discussion}\label{sec12}

Our work offers cautionary optimism on the potential role of reinforcement learning in optimizing pedagogical instructional policies. The personalized narrative AI guide may benefit students with the lowest pretest performance, without harming the performance of other learners. Indeed the average gain in scores for subjects with low (0-2) pretest scores was over 2 in both studies in the RL condition, which means the mean scores for such students at least doubled, in an assessment with 8 total points. Our results do not provide a definitive mechanism for this result,  though the engagement scores suggest that the control condition was not engaging for subjects with low pretest scores. For such students, the RL narrative AI guide condition yielded higher engagement, similar to those with higher pretest scores. This is likely due to the RL AI guide, not the narrative, since prior work found narrative alone, with hints, yielded no benefit over no narrative and no AI guide in a volume learning task~\cite{woz}.

Our encouraging result is consistent with limited  prior work  that  personalized computer-assisted learning software may
sometimes be similarly or only slightly more effective on average but may particularly benefit students who start with lower scores or take longer to complete problems (e.g.~\cite{shen2016reinforcement, de2021students}).
Since the RL algorithm we used aims to maximize expected (test) outcomes, if differentiation within the available pedagogical supports can increase the outcome of any subgroups (without harming the outcomes of other subgroups), the algorithm should learn from data to provide such personalization.  Our analysis did not find a significant benefit of RL over the control condition at the population level, though it is possible an effect would be observed with a larger sample size, or with different state feature representations, network architectures, or RL algorithms. 

Across study 1 and 2 the comparison between the narrative RL condition and control conditions appear largely stable (Figure~\ref{rlbot:fig:improve_by_pre_study}), with a trend for the RL condition benefiting those with low pretest scores,%
This suggests an RL decision policy learned on one population can sometimes benefit other populations. 

\section{Conclusion}\label{sec13}

Our work was conducted on around 400 students, which is typically less than the number of third to fifth-graders in a school district, suggesting the feasibility of using this approach to quickly optimize digital learning environments. 
By combining reinforcement learning with explainable AI, this approach can provide new insights into the interaction of context and student learning that may prompt new research in learning sciences and has a high potential to help quickly identify and scale effective learning practices. 

\backmatter

\begin{appendices}

\section{}

\subsection*{A Real Conversation Example}
To provide an indication of how well the RL AI guide performs after about 200 rounds of iterations, we present below a real conversation snippet between a user and the RL AI guide.

\begin{itemize}
    \item[] \textbf{Child}: Let's keep thinking.
    \item[] \textbf{AI guide}: [Direct Hint] You can fit 3 chocolates along the height, 5 chocolates along the width, and 6 chocolates along the length of the box.
    \item[] \textbf{Child}: Ok.
    \item[] \textbf{AI guide}: [Guided Prompt] Hm... I suggest thinking about the concept of a unit cube and the volume of a unit cube. It may remind you of what you need in order to calculate the volume of the box!
    \item[] \textit{Child inputs 90 (correct answer) for the volume question.}
    \item[] \textbf{Child}: We got another one right!
    \item[] \textbf{AI guide}: [Acknowledgement] :)
    \item[] \textbf{Child}: Do you think we multiply or divide.
    \item[] \textbf{AI guide}: [Encouragement] I think I have some ideas! But I won't spoil it. Just let me know if you need a hint.
    \item[] \textit{Child inputs 180 (correct answer) for the weight question.}
    \item[] \textbf{Child}: Also we got another one right!
    \item[] \textbf{AI guide}: [Acknowledgement] :)
\end{itemize}

As can be seen from this example, the AI guide adopted all four types of teaching strategies: direct hint, guided prompt, encouragement, and acknowledgment. In particular, when the child asked ``Do you think we multiply or divide.'', instead of providing any meaningful help, the AI guide simply encouraged the child. This turns out to be an effective teaching strategy since the child successfully solved the problem without needing specific math-related help from the AI guide.

\subsection*{The math learning activity and narrative}

The central task of the learning platform is the completion of an online math learning activity focused on teaching 2 concepts related to volume and its knowledge components (measurement and multiplication). Similar to the task in Ruan et al.~\cite{woz}, our math learning activity presented children with 6 tasks in order of increasing difficulty. All of these tasks were subparts of one overall problem (calculating if a box of chocolates weighed $>$ 320 oz) and results from prior tasks were used in later ones.

The first three tasks ask children to measure the length, width, and height of an on-screen box by dragging a bar of single-unit square chocolate along its edges. For the fourth task, children are asked to use these measurements to calculate the total number of chocolates that can fit in the box. In the fifth task, children are told that each chocolate weighs 0.5oz., the information they are required to use to help them calculate the total weight of the box. Finally, the sixth task asks children to determine if the box can be safely transported by a boat with a weight limit of 320oz.

Our AI guide support component replaces the remote human feedback support component used by Ruan et. al~\cite{woz}. In addition, due to the constraints of the covid-19 pandemic situation at the time, children completed our math learning activity remotely through an online web app as opposed to in a physical lab setting. This means children complete the online activity asynchronously without the observation or interference of a researcher. We conducted a 10-minute video call with each guardian-child to confirm there was a child learner who intended to complete the task. During this video call, we emphasized to the subjects that they should complete the activity without the help of any outside resources, and guardians were asked to ensure their children completed the task without outside resources.

\subsection*{AI guide support}

During the math learning activity, each time the AI guide is sent a message, it can take one of several actions. 1) Provide an instructional hint. Hints are specific to the task the child is currently working on and are provided in a fixed order. Each time this action is taken, the next hint is provided, and when no hints are left for the current task, the AI guide sends an appropriate message. 2) Send acknowledgment. In this case, the agent decides that no action is appropriate; the AI guide acknowledges the child’s message but otherwise provides no assistance or encouragement (“:)”). 3) Send encouragement. A random encouraging message from a predetermined list is sent to the user (for example, “You’re doing a great job. If we keep working like this, we’ll be done in no time!”). These messages were written to promote a growth mindset and excitement about the challenge of the problem without giving help to the problem itself. 4) Guided Prompt. As with normal hints, guided prompts are specific to the current task and are provided in a fixed order. In contrast to normal hints, the goal of guided prompts is to provide some assistance to children who do not need as much help as a standard hint provides (for example, “Try thinking about the concept of volume to solve this problem.”). 

The AI guide only responds when spoken to with the exception of periodic “reminder” messages which remind the children that the AI guide is there (for example, “I think you’ve got this. But if you need help, just let me know!”). These messages are chosen randomly from a predetermined list. The goal is to provide children with social support as well as remind children to use the AI guide as a helpful resource if they become stuck. These reminders are sent every 120 seconds after user inactivity (including both speaking to the AI guide and interacting with the software). Additionally, the AI guide has a list of predicted responses that it ignores (such as “Okay”) or acknowledges with “You’re welcome!” (such as “Thanks”) to reduce noise from natural language responses that do not require one of the above actions.

In contrast to the experimental condition, there was no AI guide and no hint system present in the control condition.

The AI guide responds to input from the learner. There was an automated reminder for the child to engage if no prior interactions had happened during the 120 seconds. The automated instructional policy was trained using reinforcement learning. For the first phase, the reward model uses $\alpha = 0.01, \beta = 0.1, \gamma = 0.3$, based on the hyperparameter choices of prior work~\cite{bassen2020reinforcement} and our earlier simulations simulation. 

\begin{figure}[!th]%
    \centering
    \includegraphics[width=0.9\linewidth]{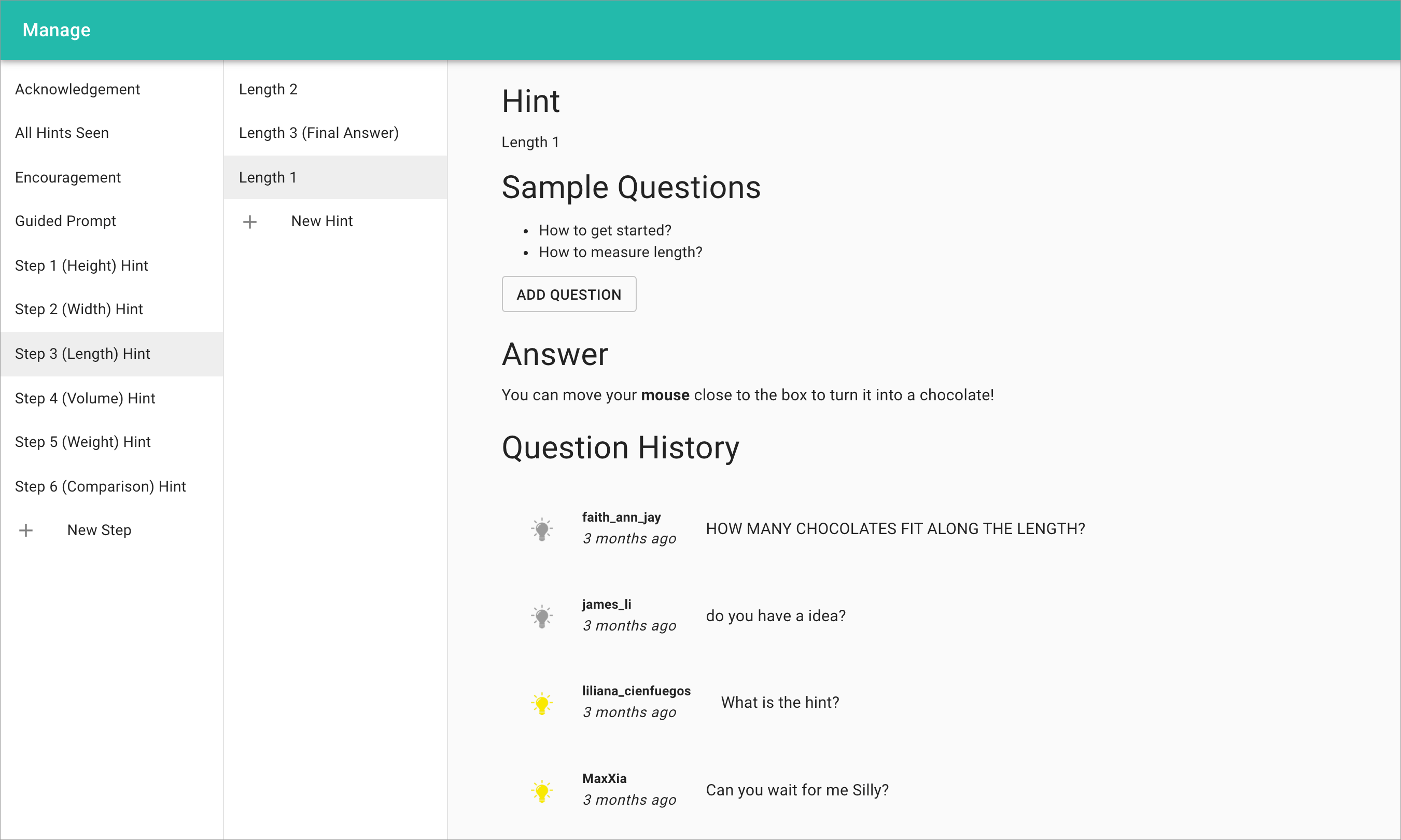}
    \caption{Interface for teachers to write hints and prompts.}
    \label{rlbot:fig:dashboard}
\end{figure}

All the hints and message templates were written and uploaded through an easy-to-use teacher-facing dashboard (see Figure \ref{rlbot:fig:dashboard}) by educators and designers without prior background in machine learning. 

\subsection*{Distribution of Grade and Pre-test Scores in Treatment and Control}

\textbf{In study 1}, 339 participants in grades 3–5 were recruited through Twitter, NextDoor, userinterview.com, school mailing lists, and word of mouth. Children came from 263 different schools. Of 339 participating children, 172 were boys and 167 were girls. 114 were in grade 3, 114 were in grade 4, and 111 were in grade 5.  Children were randomly assigned to one of the two systems based on a predetermined ratio: 70 children used the control system and 269 children used the system with RL AI guide-mediated guidance. Gender and grade were balanced across the two conditions.

There is no significant difference between the treatment and control group in study 1 on pre-test score (Cohen's d $-0.235$, two-sample Wilcoxon rank test $W = 10782$, p-value $= 0.058$) as well as grade (Cohen's d $0.008$, two-sample Wilcoxon rank test $W = 9371.5$, p-value $= 0.9502$).

\textbf{In study 2}, 35 participants were recruited using userinterview.com and  childrenhelpingscience.com. There is no significant difference between the treatment and control group in study 2 on pre-test score (Cohen's d $-0.262$, two-sample Wilcoxon rank test $W = 175.5$, p-value $= 0.4634$), as well as grade (Cohen's d $0.282$, two-sample Wilcoxon rank test $W = 136.5$, p-value $= 0.5665$).

\subsection*{Details on Repeated Post-test Taking in Logged Data}

The software did not explicitly check for students repeating the pretest or post-test, and in our post-analysis, we found a few students took either the pretest or post-test test multiple times. The logging software only recorded the score of the final time the student took the test. For this reason, we only analyzed students who took the pretest and post-test once. In study 1, this resulted in 68 (out of 70) students in the control condition being kept in the analysis (only 2 students took either the pretest or posttest twice) and 258 (out of 269) students in the RL condition. In study 2, 18 (out of 19) students in the control condition, and 17 (out of 18) students in the RL condition were included in the analysis. We computed our results after removing these duplicate entries.

\subsection*{Report on Time Spent Between Control and RL Condition}

On average students do often spend longer\footnote{We excluded individuals who took longer than 90 minutes on the task in this figure since such subjects are likely to have taken breaks. All individuals who took at least 90 minutes took over 2 hours, and there were 8 such individuals excluded using this restriction.}  on the RL narrative condition task than in the control condition: Figure~\ref{rlbot:fig:time_on_task}. This was consistent for students across all three groups of pretest performance, and the difference in time spent between the two conditions was largely similar for all three groups. As it was only students in the low pretest group that seem to have a significant benefit from the RL condition, it seems unlikely that time on task is the primary reason for improved performance in the RL narrative condition. 

We report the time spent on the pretest, task, and post-test (assessment), in each control and experiment, in both study 1 and study 2 (see Table~\ref{rlbot:tab:time-spent}). We conduct a two-sample Wilcoxon rank test on all pairs (between study 1 and study 2). We find no significance between the two studies.

\begin{table*}[ht]
\centering
\caption{Average time (secs) spent on different parts of the tutoring session.}
\vspace{1mm}
\begin{tabular}{@{}r|cc|cc@{}}
\toprule
\multicolumn{1}{l}{} & \multicolumn{2}{l}{Online RL Study 1} & \multicolumn{2}{l}{Distilled Policy Study 2} \\ 
\multicolumn{1}{l}{} & Control         & Narrative AI        & Control            & Narrative AI            \\ \midrule
Pre-test             & 410.5           & 424.3               & 478.2              & 560.5                   \\
Task                 & 592.3           & 2334.8              & 542.1              & 1978.9                  \\
Post-test            & 511.1           & 290.7               & 246.5              &  300.9                  \\ \bottomrule
\end{tabular}
\label{rlbot:tab:time-spent}
\end{table*}

\subsection*{Engagement}
In study 1, students with low pretest scores (scores 0-2) had an average engagement score of 2.67 (N=14, standard error = 0.23) in the control condition and an average engagement score of 3.29 (N=40, standard error = 0.11) in the RL narrative AI guide condition. In study 1, students with medium pretest scores (scores 3-5) had an average engagement score of 3.43 (N=15, standard error = 0.12) in the control condition, and an average engagement score of 3.36 (N=105, standard error = 0.05) in the RL narrative AI guide condition. In study 1, students with high pretest scores (scores 6-8) had an average engagement score of 3.24 (N=38, standard error = 0.07) in the control condition, and an average engagement score of 3.48 (N=108, standard error = 0.04) in the RL narrative AI guide condition. Three subjects in study 1 did not complete the engagement survey. 

In study 2, students with low pretest scores (scores 0-2) had an average engagement score of 2.71 (N=5, standard error = 0.12) in the control condition and an average engagement score of 3.28 (N=7, standard error = 0.23) in the RL narrative AI guide condition. In study 2, students with medium pretest scores (scores 3-5) had an average engagement score of 3.21 (N=6, standard error = 0.23) in the control condition, and an average engagement score of 3.28 (N=6, standard error = 0.28) in the RL narrative AI guide condition. In study 2, students with high pretest scores (scores 6-8) had an average engagement score of 3.45 (N=7, standard error = 0.17) in the control condition, and an average engagement score of 3.29 (N=4, standard error = 0.23) in the RL narrative AI guide condition.

\subsection*{Implementation Details}

\begin{figure}[!th]%
    \centering
    \includegraphics[width=0.9\linewidth]{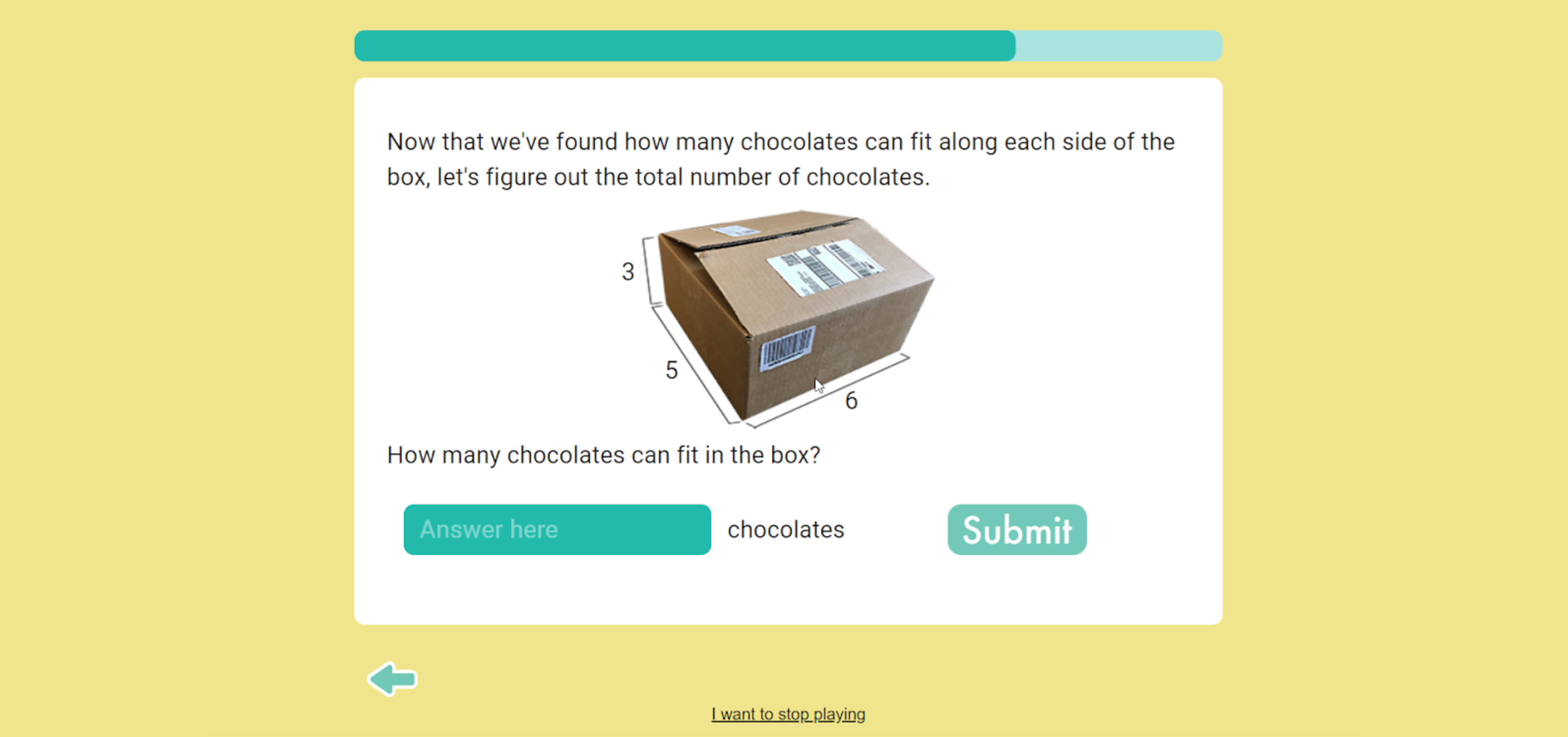} 
    \caption{The control interface where the child attempts to solve the math problem without any narrative or help from a chatbot. The child can click on ``I want to stop playing'' to quit the activity at any time.}
    \label{rlbot:fig:control_interface}
\end{figure}

The platform consists of three major parts: a user-facing interactive website (Figure \ref{rlbot:fig:control_interface} for control and Figure \ref{rlbot:fig:chatbot_interface} for AI guide), an admin dashboard (Figure \ref{rlbot:fig:dashboard}), and an AI guide server. Both the website and the dashboard were created using Web technologies, including ReactJS \cite{react} and TypeScript \cite{typescript}. The Python-based AI guide was hosted on an AWS server and used Flask \cite{flask} as its API gateway to expose essential functions. The interactive website communicated with a GraphQL \cite{graphql} API endpoint backed by Hasura Engine \cite{graphql} and PostgreSQL \cite{postgres}. The stored user conversation data was reflected in real-time on the admin console, where researchers could view the chat history and modify message templates. All user data was uploaded to the backend by Google App Script upon the completion of the user's session. 

Questionnaires and quizzes were created using Google Forms, and we used HTML iframe to embed Google Forms into the website to automatically process the form responses so as to enable real-time RL. When users interact with the AI guide, the observation space is calculated in real-time, and the AI guide performs action selection to reply to users. When users completed the post-quiz, their answers were converted to vector inputs and fed into the RL AI guide in real-time, which triggered a webhook to request the AI guide server to update its model accordingly. A complete diagram showing the interaction between the user and the RL agent is displayed in Figure \ref{rlbot:fig:rl_diagram}.

\subsection*{Authors' contributions}

S.R., A.N., W.S., J.H. J.Z., M.G., Y.L, K N. C.W., R.Y. J.L, and E.B. conducted  research. S.R., A.N., W.S.,Y.L., J.L., and E.B. designed research. S.R., A.N.,W.S., and E.B. performed the analysis. S.R., W.S., A.N., and E.B. wrote  the manuscript. 

\noindent S.R.(Author One) contributed equally to this work with A.N. (Author Two).

\noindent E.B. is to whom correspondence should be addressed. E-mail: ebrun@stanford.edu

\end{appendices}

\bibliography{sn-bibliography}%

\end{document}